\documentclass[5p, times]{elsarticle}
\makeatletter
\def\ps@pprintTitle{%
 \let\@oddhead\@empty
 \let\@evenhead\@empty
 \def\@oddfoot{Published in Neurocomputing, \url{https://doi.org/10.1016/j.neucom.2024.128918}\hfill}%
 \let\@evenfoot\@oddfoot}
\makeatother

\usepackage{graphicx}
\usepackage{amssymb,amsmath,array}
\usepackage[ruled,vlined,linesnumbered]{algorithm2e}
\usepackage{booktabs}
\usepackage[table,xcdraw]{xcolor}
\usepackage[colorlinks]{hyperref}
\usepackage[nameinlink,capitalize]{cleveref}
\usepackage[symbol]{footmisc}
\usepackage{soul}
\usepackage{xfrac}
\usepackage{multirow}
\usepackage{adjustbox}
\usepackage{courier}
\usepackage{caption}
\usepackage{subcaption}
\usepackage{graphicx}

\usepackage{siunitx}
\newrobustcmd\B{\DeclareFontSeriesDefault[rm]{bf}{b}\bfseries}

\newcommand{\numclasses}{\ensuremath{Y}\xspace}
\newcommand{\classindex}{\ensuremath{y}\xspace}

\newcommand{\steps}{\ensuremath{K}\xspace}
\newcommand{\stepindex}{\ensuremath{k}\xspace}

\newcommand{\model}{\ensuremath{M}\xspace}
\newcommand{\modelindex}{\ensuremath{m}\xspace}
\newcommand{\loss}{\ensuremath{L}\xspace}
\newcommand{\losses}{\ensuremath{\mathbb L}\xspace}
\newcommand{\optimizer}{\ensuremath{u}\xspace}
\newcommand{\optimizers}{\ensuremath{\mathbb U}\xspace}
\newcommand{\scheduler}{\ensuremath{s}\xspace}
\newcommand{\schedulers}{\ensuremath{\mathbb S}\xspace}
\newcommand{\configuration}{\ensuremath{C}\xspace}
\newcommand{\configurations}{\ensuremath{\mathbb C}\xspace}
\newcommand{\hyperparameter}{\ensuremath{h}\xspace}
\newcommand{\hyperparameters}{\ensuremath{\mathbb H}\xspace}

\newcommand{\numinit}{\ensuremath{P}\xspace}

\newcommand{\numtrials}{\ensuremath{T}\xspace}

\newcommand{\learningrate}{\ensuremath{\gamma}\xspace}

\newcommand{\momentum}{\ensuremath{\mu}\xspace}

\newcommand{\robustaccuracy}{\text{RA}\xspace}

\newcommand{\myparagraph}[1]{\noindent \textbf{#1}}

\newcommand{\vct}[1]{\ensuremath{\boldsymbol{#1}}}

\newcommand{\set}[1]{\ensuremath{\mathcal{#1}}}

\newcommand{\argmax}{\operatornamewithlimits{\arg\,\max}}
\newcommand{\med}{\text{med}\xspace}

\newcommand{\argmin}{\operatornamewithlimits{\arg\,\min}}

\newcommand{\ie}{{i.e.}\xspace}
\newcommand{\eg}{{e.g.}\xspace}

\newcommand{\ellone}{$\ell_1$\xspace}
\newcommand{\elltwo}{$\ell_2$\xspace}
\newcommand{\ellinf}{$\ell_{\infty}$\xspace}

\newcommand{\sgd}{\texttt{GD}\xspace}
\newcommand{\adam}{\texttt{Adam}\xspace}
\newcommand{\adamax}{\texttt{AdaMax}\xspace}

\newcommand{\logitloss}{\texttt{LL}\xspace}
\newcommand{\ce}{\texttt{CE}\xspace}
\newcommand{\dlr}{\texttt{DLR}\xspace}

\newcommand{\calr}{\texttt{CALR}\xspace}
\newcommand{\rlrop}{\texttt{RLRoP}\xspace}
\newcommand{\fixed}{\texttt{Fixed}\xspace}

\DeclareMathOperator*{\subjectto}{s.t.}

\journal{Neurocomputing}

\begin{document}

\begin{frontmatter}

\title{HO-FMN: Hyperparameter Optimization for Fast Minimum-Norm Attacks}

\author[UNICA]{Raffaele Mura\footnote[1]{These authors contributed equally to this work.}}
\author[UNICA]{Giuseppe Floris$^*$}
\author[UNICA,SAPIENZA]{Luca Scionis$^*$}
\author[UNICA,SAPIENZA]{Giorgio Piras}
\author[UNICA]{Maura Pintor\footnote[2]{Corresponding Author. Email: \texttt{maura.pintor@unica.it}}}
\author[UNICA]{Ambra Demontis}
\author[UNICA]{Giorgio Giacinto}
\author[UNICA]{Battista Biggio}
\author[UNIGE,UNICA]{Fabio Roli}

\affiliation[UNICA]{
organization={University of Cagliari, Dept. of Electrical and Electronic Engineering},
city={Cagliari},
country={Italy}}
\affiliation[SAPIENZA]{
organization={Sapienza University of Rome, Department of Computer Engineering},
city={Rome},
country={Italy}}
\affiliation[UNIGE]{
organization={University of Genoa, Department of Computer Science, Bioengineering, Robotics and Systems Engineering},
city={Genoa},
country={Italy}}

\begin{abstract}
Gradient-based attacks are a primary tool to evaluate robustness of machine-learning models. However, many attacks tend to provide overly-optimistic evaluations as they use fixed loss functions, optimizers, step-size schedulers, and default hyperparameters.
In this work, we tackle these limitations by 
proposing a parametric variation of the well-known fast minimum-norm attack algorithm, whose loss, optimizer, step-size scheduler, and hyperparameters can be dynamically adjusted. 
We re-evaluate 12 robust models, showing that our attack finds  smaller adversarial perturbations without requiring any additional tuning. This also enables reporting adversarial robustness as a function of the perturbation budget, providing a more complete evaluation than that offered by fixed-budget attacks, while remaining efficient. 
We release our open-source code at \url{https://github.com/pralab/HO-FMN}.
\end{abstract}




\end{frontmatter}

\section{Introduction} \label{sec:intro}
Machine learning (ML) models are susceptible to adversarial examples~\cite{biggio13-ecml, szegedy2014intriguing}, i.e., input samples that are intentionally perturbed to mislead the model. 
Such samples are optimized using gradient-based attacks, which allow one to efficiently find adversarial perturbations close enough to the original unperturbed samples.
However, using gradient-based attacks can only provide an \textit{empirical} estimation of adversarial robustness. In particular, if an attack fails to find an adversarial example, we cannot prove that the given input is robust (\ie there are adversarial manipulations that make the input adversarial, but the attack was not able to find it), similarly to what happens when searching for bugs in software.
Ideally, through an exhaustive search or formal verification procedure, it would be possible to provide a guaranteed robustness assessment, which is however practically infeasible due to the high computational requirements and dimensionality entailing the problem~\cite{kumar2020curse}.
This means that gradient-based attacks are most likely to provide an overly-optimistic estimation of adversarial robustness, and obtaining more reliable evaluations is not trivial~\cite{pintor22_indicators}. 
It has been indeed shown that several defenses proposed to improve robustness to adversarial examples were wrongly evaluated, and rather than improving adversarial robustness they were simply \textit{obfuscating gradients}, thereby invalidating the optimization of gradient-based attacks~\cite{carlini2017towards,carlini2019evaluating,tramer2020adaptive,pintor22_indicators}.
This problem is also exacerbated by the fact that each attack presents different hyperparameters which require careful tuning to be executed correctly, i.e., to find a better optimum, along with fixed choices for the loss function, optimizer, and step-size scheduling algorithm. Nevertheless, in many of the reported evaluations, such attacks are run with their default settings, even if it has been shown that this may result in overestimating adversarial robustness \cite{carlini2019evaluating, pintor22_indicators,tramer2020adaptive}.
Finally, many robustness evaluations are obtained by running fixed-budget attacks that only provide the adversarial robustness estimate at a fixed perturbation budget $\epsilon$, without providing any insight on the robustness of the model when adversarial perturbations have a different size.
To summarize, the main problems hindering the diffusion of more reliable adversarial robustness evaluations are:
    (i) the use of fixed loss, optimizer, and step-size scheduler, along with default attack hyperparameters; and
    (ii) the use of fixed-budget attacks, providing only limited insights on how models withstand adversarial attacks.

\begin{figure*}[h]
    \centering
\includegraphics[width=0.8\textwidth]{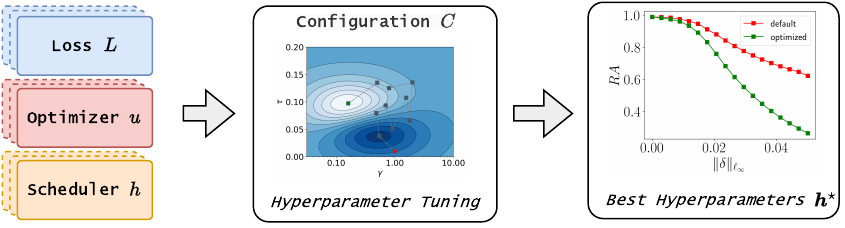}
    \caption{Overview of our HO-FMN approach.}
    \label{fig:overview}
\end{figure*}

In this work, we aim to overcome these limitations by improving the current version of the Fast Minimum-Norm (FMN) attack originally proposed in~\cite{pintor2021fast}. 
To this end, we first propose a modular reformulation of the FMN attack that enables the use of different loss functions, optimizers, and step-size schedulers (\autoref{sec:fmn}). This facilitates the task of finding the strongest FMN configuration against each given model.
We then leverage Bayesian optimization to perform a hyperparameter-optimization step that, for any given FMN configuration, automatically finds the best hyperparameters for the optimizer and the scheduler of choice (\autoref{sec:ho-fmn}). An overview of our method, referred to as Hyperparameter Optimization for Fast Minimum-Norm (HO-FMN) attacks, is presented in \autoref{fig:overview}.
We extensively evaluate HO-FMN on 12 robust models against competing baseline attacks, supporting the validity of our method on efficiently obtaining complete robustness evaluation curves of ML models (\autoref{sec:experiments}).
With respect to our preliminary work in~\cite{esann23-autofmn}, we extend the current approach by revisiting the FMN attack algorithm and rethinking the hyperparameter optimization framework with a Bayesian approach. 
In addition, we expand our experimental setup to get a more accurate evaluation. 
We conclude by discussing related work (\autoref{sec:related}), along with the main contributions, limitations, and future research directions (\autoref{sec:conclusions}).

\section{Revisiting Fast Minimum-Norm Attacks}\label{sec:fmn}

We first present the FMN attack as originally proposed in~\cite{pintor2021fast}, highlighting the changes applied to obtain the modular version of the attack algorithm in which we parameterize the loss function, the optimizer, and the step-size scheduler, along with their hyperparameters. 
The proposed reformulation of the algorithm enables the selection of each component independently, creating multiple parametric variations of the original attack.

\myparagraph{Notation.}
Our goal is to discover minimum-norm adversarial perturbations that cause a model to misclassify an input. 
Let
$\vct x \in \set X = [0,1]^d$ represent a $d$-dimensional input data point
with true label $y \in \set Y = \{1, \ldots, \numclasses\}$. 
We denote the target
function as $f: \set X \times \Theta \mapsto \set Y$, where $\vct \theta
\in \Theta$ is its set of parameters. 
We
will utilize $f$ for the label prediction function and $f_\classindex$ to refer to the
continuous output (logit) corresponding to each class $\classindex \in \set Y$, .

\myparagraph{Attack Formulation.} Minimum-norm attacks aim to find the smallest perturbation possible $\vct \delta^\star$ for which a sample $\vct x$ labeled as $y$ gets misclassified by a model with parameters $\vct \theta$, \ie $f(\vct x + \vct \delta^\star; \vct \theta) = \argmax_{\classindex \in 1, ... \numclasses} f_\classindex(\vct x + \vct \delta^\star; \vct \theta) \neq y$. 
Their goal is thus to solve the following optimization problem: 
\begin{align}
	\label{eq:obj} \vct \delta^\star \in \quad & \argmin_{\vct \delta}  \|\vct \delta \|_{p} \, ,\\
\label{eq:constr} \subjectto \quad &  \loss_\logitloss(\vct x + \vct \delta, y, \vct \theta) < 0 \, , \\
\label{eq:bounds}	& \vct x + \vct \delta \in [0,1]^d \, ,
\end{align}
where $\| \cdot \|_p$ indicates the chosen $\ell_p$ norm ($p=1,2,\infty$). The constraint in \cref{eq:constr} is the difference-of-logits (\logitloss) loss, defined as:
\begin{equation}
\label{eq:logitloss}
\loss_\logitloss(\vct x, y, \vct \theta) =  f_{y}(\vct x, \vct \theta) - \max_{j \neq y}f_{j}(\vct x, \vct \theta) \, .
\end{equation}
This loss is negative when the input is misclassified. The box constraint in \cref{eq:bounds} ensures that the sample $\vct x + \vct \delta$ remains in the feasible input space.

The Fast Minimum-Norm (FMN) attack~\cite{pintor2021fast} proposes a reformulation of the minimization problem to find the smallest $\epsilon$ for which the constraint is satisfied. The problem is reformulated as follows:
\begin{align}
\label{eq:constr-eps}
\min_{\epsilon, \vct \delta} \,  \epsilon \, , \; \; \;  {\rm s.t.} \, \| \vct \delta \|_p \leq \epsilon, \,
\end{align}
plus the constraints in Eqs.~\eqref{eq:constr}-\eqref{eq:bounds}, where $\epsilon$ is an upper bound on the perturbation size $\| \vct \delta\|_p$.
We now discuss the algorithm used by FMN.

\begin{algorithm}[t]
 \SetKwInOut{Input}{Input}
    \SetKwInOut{Output}{Output}
    \SetKwComment{Comment}{$\triangleright$\ }{}
    \DontPrintSemicolon
	\caption{Fast Minimum-Norm (FMN) Attack Algorithm.}
	\label{alg:modular_fmn}
   \Input{$\vct x$, the input sample; $y$, the target (true) class label; $\alpha_{0}$, the initial $\vct \delta$-step size; \steps, the total number of iterations; \textcolor{blue}{\loss, the attack loss}; \textcolor{red}{\scheduler, the step size scheduler}; \textcolor{cyan}{\optimizer, the optimizer}.}
	\Output{The minimum-norm adversarial example $\vct x^\star$.}
    $\epsilon_0 = \infty, \, \vct \delta_{0} \gets \vct 0, \, \vct \delta^\star \gets \vct \infty$, $\gamma_{0}=0.05$ \label{line:init}\Comment*[r]{initialization}
    \For {$\stepindex = 1, \ldots, \steps$} 
        {
        $\gamma_{\stepindex} \gets \scheduler_\gamma(\gamma_{0}, \stepindex, \steps)$ \Comment*[r]{$\epsilon$-step size decay} \label{line:gamma_decay}
        $\epsilon_{\stepindex} = \optimizer_\epsilon(\epsilon_{\stepindex-1}, \gamma_\stepindex, \|\vct \delta_\stepindex\|_p)$\label{line:eps_step} \Comment*[r]{$\epsilon$-step}
	    $\vct g \gets \nabla_{\vct \delta} \textcolor{blue}{\loss}(\vct x +\vct \delta_{\stepindex-1}, y,  \vct \theta)$ \Comment*[r]{\textcolor{blue}{loss} gradient}\label{line:gradient}
        $\alpha_{\stepindex} \gets \textcolor{red}{\scheduler}(\alpha_{0}, \stepindex, \steps)$ \Comment*[r]{\textcolor{red}{scheduler} step}\label{line:step_decay}
        $\vct \delta_{\stepindex} \gets$ \textcolor{cyan}{\optimizer}($\vct \delta_{\stepindex-1}, \texttt{proj}(\vct g) , \alpha_{\stepindex}$) \label{line:opt_step} \Comment*[r]{\textcolor{cyan}{optimizer} $\vct \delta$-step}
        $\vct \delta_\stepindex \gets \Pi (\vct x, \vct \delta_\stepindex, \epsilon_\stepindex)$ \label{line:project} \Comment*[r]{proj. onto feasible domain}
		}
	\textbf{return} $\vct x^\star \gets \vct x + \texttt{best}(\vct \delta_0, ... \vct \delta_\steps$) \label{line:return} \Comment*[r]{return best solution}
\end{algorithm}

\myparagraph{FMN Attack Algorithm.} We report in \autoref{alg:modular_fmn} a revisited formulation of the FMN attack, in which we emphasize our specific contributions to make it parametric to its components. 
First, the attack is initialized (\autoref{line:init}), where the initial perturbation is set to $\vct \delta = \vct 0$ and the initial constraint is set to $\epsilon_0 = \infty$ to encourage the initial exploration of the loss landscape without encountering constraints in this phase.\footnote{We simplify the algorithm by removing a refined estimate of $\epsilon_0$ that approximates the distance to the boundary using the gradient of the loss used by FMN, as it might require computing an additional gradient when using our general algorithm with different losses.}
Then, the original FMN algorithm develops as a two-step process: the $\epsilon$-step minimizes the upper bound constraint on the maximum perturbation by reducing $\epsilon$ as long as the sample is adversarial, and the $\vct \delta$-step updates the perturbation towards the adversarial region trying to find it within the constraint defined by $\epsilon$. 
The $\epsilon$-step is controlled by a parameter $\gamma_\stepindex$ that modifies the multiplicative factor for the current $\epsilon$.
The parameter $\gamma_\stepindex$, in turn, is reduced with cosine annealing decay (\autoref{line:gamma_decay}).  
At each iteration, $\epsilon$ is reduced (increased) by a factor $1 - \gamma_\stepindex$ (by a factor $1 + \gamma_\stepindex$) if $\vct x + \vct \delta$ is (not) adversarial (\autoref{line:eps_step}).
Subsequently, the $\vct \delta$-step updates the perturbation with the gradient of the loss function $\loss(\vct x, y, \vct \theta)$ (\autoref{line:gradient}).
While FMN uses the Logit Loss (\logitloss) of \cref{eq:logitloss}, we modify the algorithm to work with any differentiable loss \loss.

The FMN attack normalizes the gradients $\nabla_x \loss(\vct x, y, \vct \theta)$ in the \elltwo norm, \ie $\vct g^\prime = \sfrac{\vct g}{\|\vct g\|_2}$, and multiplies it by a step size $\alpha_k$.
In our formulation, we generalize this step with a linear projection (\texttt{proj}) onto a unitary-sized $\ell_p$-ball.
The projection maximizes a linear approximation of the gradient within a unitary $\ell_p$-ball, as $\vct g^\prime =\underset{\|\vct v\|_p \leq 1}{\argmax} \; \vct v^\top \vct g$.
This is accomplished, in \ellinf, by taking the sign of the gradient $\texttt{sign}(\vct g)$, and produces a dense update of all components of $\vct \delta_\stepindex$.
Without loss of generality, the projection can be achieved in \ellone and \elltwo by changing the norm used in the maximization.

In FMN, the step size is decayed with a decay schedule rule (\autoref{line:step_decay}).
In the original formulation, the decay was regulated with a Cosine Annealing Learning Rate scheduler (\calr). 
Our algorithm makes the scheduler \scheduler parametric, unlocking new scheduler rules for tuning the step size. 

The perturbation is then updated with the computed $\vct \delta$-step (\autoref{line:opt_step}), where we modify the original gradient descent (\sgd), replacing it with a generic optimizer \optimizer that can use different algorithms, \eg momentum strategies.
The perturbation is then projected onto the $\epsilon$-ball and clipped to maintain the modified sample within the input domain (\autoref{line:project}).
Finally, the best perturbation is returned (\autoref{line:return}).

\myparagraph{Summary of Changes from FMN.} While the overall algorithm remains conceptually unchanged, we adjust the attack loss \loss, the optimizer \optimizer, and the step-size scheduler \scheduler used in the $\vct \delta$-step to make them interchangeable. 
These elements were fixed in the original attack implementation, while in our work, we make them general and allow multiple choices for each component. 
The generalization of the loss \loss required an additional modification to the original algorithm, where the $\epsilon$-step size was computed in the initial steps by estimating the distance to the boundary to speed up convergence. 
This estimation required computing the gradient of the \logitloss loss, thus would require an additional backward pass for each iteration. 
With preliminary experiments, we found that such estimation improves the query efficiency of the initial steps, but it does not change substantially the final outcome of the attack.
In addition, we use the linear projection of the gradient $\vct g$ instead of the normalization.
Most significantly, our changes to the FMN algorithm required a thorough reevaluation of its implementation. 
Specifically, we enabled the choice of losses, optimizers, and schedulers that were already available in widely-adopted deep learning frameworks, which are commonly used (and efficiently implemented) for training deep neural networks.

\myparagraph{Why FMN.} Contrary to fixed-budget attacks (such as PGD~\cite{madry2017towards} and AA~\cite{croce2020autoattack}), FMN finds minimum-norm adversarial examples, solving the optimization problem in \cref{eq:obj}. 
It follows that, instead of having a scalar robustness evaluation associated with a predefined perturbation budget $\epsilon$ from a single run, we can obtain an entire curve, which we denote as a \textit{robustness evaluation curve}~\cite{biggio18}, plotting how the robust accuracy of a model decreases as the perturbation budget $\epsilon$ is increased. 
The curve can be computed efficiently from the minimum distances $\|\vct \delta^\star\|_p$ returned by FMN, by computing:
\begin{equation}
    \frac{1}{|\mathcal{D}|}\sum_{(\vct x, y) \in \mathcal{D}} \mathbb{I}(\|\vct \delta^\star\|_p<\epsilon \wedge f(\vct x + \vct \delta^\star) \neq y) \, ,
\end{equation}
for increasing values of $\epsilon$, where $\mathbb{I}$ is the indicator function, which returns $1$ ($0$) if the argument is true (false). 
These curves enable a more complete and informative robustness evaluation than that provided for a fixed $\epsilon$ value~\cite{carlini2019evaluating}.

\section{Hyperparameter Optimization for Fast Minimum-Norm Attacks}
\label{sec:ho-fmn}

A graphical representation of our HO-FMN method is depicted in \autoref{fig:overview}.
By leveraging the modular version of FMN presented in \autoref{sec:fmn}, and selecting a pool of losses, optimizers, and step-size schedulers, we (i) create multiple configurations of the FMN attack, (ii) optimize the hyperparameters of each configuration through Bayesian optimization, and rank them based on their effectiveness against the model under test, and (iii) run the attack with the best configurations found to estimate the robustness of the model. 
We discuss the choice of the configurations in \autoref{sec:ho-fmn_configs}, and, in \autoref{sec:ho-fmn_optim}, the subsequent optimization framework to get the best hyperparameters, compactly represented also in \autoref{alg:ho_fmn}. 

\subsection{HO-FMN: Configurations and Hyperparameters}\label{sec:ho-fmn_configs}
In our modular FMN re-implementation, we define the loss~\loss, the optimizer~\optimizer, and the step-size scheduler~\scheduler as parametric. 
Accordingly, by defining a pool of losses \losses, optimizers \optimizers, and schedulers \schedulers, we create diverse FMN  configurations,  each represented as a tuple $\configuration = (\loss, \optimizer, \scheduler)$. 
We then define the hyperparameter search space by associating to each configuration $\configuration$ the set of hyperparameters $\vct \hyperparameter = \vct \hyperparameter_\optimizer \cup \vct \hyperparameter_\scheduler$ of the corresponding optimizer \optimizer and scheduler \scheduler composing the final attack (\autoref{sec:ho-fmn_optim}). 

Next, given an input model, we optimize $N$ configurations $\configuration_1, \configuration_2, ..., \configuration_N \in \configurations$ to find the best set of hyperparameters for each (\autoref{sec:ho-fmn_optim}).

\myparagraph{Configuration Set.} We denote the set of attack configurations as
$\configurations := \{\configuration_1, \configuration_2,..., \configuration_N\} \,$, 
 each represented as $(L, u, s)$.
The total number $N$ of possible configurations is thus obtained as the Cartesian product of each set, i.e., $N = |\configurations| = |\losses| \times |\optimizers| \times |\schedulers|$. 
Each configuration $\configuration_i$ corresponds to the modular version of the FMN attack using a specific loss $\loss \in \losses$, optimizer $\optimizer \in \optimizers$, and scheduler $\scheduler \in \schedulers$ in its attack algorithm.
All-in-all, given a model with parameters $\vct \theta_\modelindex \in \vct \Theta$, our framework starts by considering all $N$ (or less, since not every optimizer is necessarily associated with a scheduler) configurations.

\myparagraph{Hyperparameter Search Space.}
Upon defining the configurations, the entire set \configurations undergoes a hyperparameter optimization routine.
The goal of this routine is to find, for a given target model, the best set of hyperparameters $\vct \hyperparameter_i^\star$ to be associated with each configuration $\configuration_i$. 
Note that the best set might change from one model to another. Thus the hyperparameter optimization step has to be performed anew when a different model is selected.
The search space and dimensionality of $\vct \hyperparameter_i$ vary depending on the configuration $\configuration_i$, as the chosen optimizer and scheduler may take different (and a different number of) arguments as inputs. 
In this regard, given a  configuration $\configuration_i \in \configurations$, identified by $(L, u, s)$, its set of hyperparameters is given as $\vct \hyperparameter_i = \vct \hyperparameter_{i_\optimizer} \cup \vct \hyperparameter_{i_\scheduler}$ 
(where $\vct \hyperparameter_{i_\optimizer}$ and $\vct \hyperparameter_{i_\scheduler}$ represent, respectively, the optimizer and scheduler hyperparameters, as described above).
Hence, after creating the set \configurations, the optimization procedure aims to find the best set $\vct \hyperparameter_i^\star$ for each $\configuration_i$.

\subsection{HO-FMN: Optimization Procedure}~\label{sec:ho-fmn_optim}

\begin{algorithm}[htbp]
 \SetKwInOut{Input}{Input}
    \SetKwInOut{Output}{Output}
    \SetKwComment{Comment}{$\triangleright$\ }{}
    \DontPrintSemicolon
	\caption{HO-FMN.}
	\label{alg:ho_fmn}
   \Input{
   $\set D = (\vct x, y)$, the validation dataset;
   $\configuration$, the configuration with loss \loss, optimizer \optimizer, and scheduler \scheduler; 
   \numtrials, the number of trials; \numinit, the number of initial samples to fit the regressor.
   }
	\Output{The set of best hyperparameters $\vct \hyperparameter^\star_i$.}
        $\set S = \varnothing$, $\texttt{best\_median} = \infty$\label{line:init_set}\Comment*[r]{observation history}
        \For{$j = 1, \ldots, \numinit$} 
        { 
        $\vct \hyperparameter_{j} = \texttt{gen\_h()}$\label{line:sobol}\Comment*[r]{sample first hypers}
        $\vct x^\star_j$ = $\texttt{FMN}_{\configuration, \vct \hyperparameter_j}(\vct x, y)$ \label{line:fmn_1}\Comment*[r]{initial observations}
        $\Tilde{\|\vct \delta_j\|}$ = $\med(\|\vct x - \vct x^\star_j\|)$ \label{line:comp_med}\Comment*[r]{compute median}
            $\set S \gets \set S \cup (\vct \hyperparameter_{j}, \Tilde{\|\vct \delta_j\|})$ \label{line:update_set1}\Comment*[r]{update observations}
        }
        \For{$j = P+1, \ldots, \numtrials$} 
        {  
        $\texttt{gpr.fit}(\set S)$ \label{line:fit_reg}\Comment*[r]{fit GP regressor}
         $\vct \hyperparameter_{j}$ = $\texttt{a}(\texttt{gpr.mean}, \texttt{gpr.std})$ \label{line:acquire}\Comment*[r]{acquire new $\vct h$} 
            $\vct x^\star_j$ = $\texttt{FMN}_{\configuration, \vct \hyperparameter_j}(\vct x, y)$ \label{line:fmn2}\Comment*[r]{new observations}
            $\Tilde{\|\vct \delta_j\|}$ = $\med(\|\vct x - \vct x^\star_j\|)$ \label{line:comp_med2}\Comment*[r]{compute median}
            $\set S \gets \set S \cup (\vct \hyperparameter_{j}, \Tilde{\|\vct \delta_j\|})$\label{line:update_set2}\Comment*[r]{update observations}
            \If{$\Tilde{\|\vct \delta_j\|} <$ \texttt{best\_median }} 
            { 
            $\vct \hyperparameter_i^\star = \vct \hyperparameter_{j}$ \label{line:store_best_hypers}\Comment*[r]{store best $\vct \hyperparameter$}
            \texttt{best\_median} = $\Tilde{\|\vct \delta_j\|}$ \label{line:update_best_hypers}\Comment*[r]{update best median}
            }
        }
    \textbf{return} $\vct \hyperparameter^\star$ \label{line:return_hyper} \Comment*[r]{return best solution}
\end{algorithm}

We show in \autoref{alg:ho_fmn} the complete HO-FMN procedure, which we use to find $\vct \hyperparameter^\star_i$ for each $\configuration_i$. 
This amounts to finding the set $\vct \hyperparameter^\star$ that minimizes the median perturbation $\Tilde{\|\vct \delta\|} = \med(\|\vct x - \vct x^\star\|) $, where \med is the median function and $\vct x^\star = \texttt{FMN}_{C, \vct h}(\vct x, y)$, \ie, the output of \autoref{alg:modular_fmn} when using configuration $\configuration$ with hyperparameters $\vct \hyperparameter$. We select the median perturbation size as the objective to minimize for obtaining the best hyperparameters, following~\cite{pintor2021fast}, as it reduces the impact of potential outliers that may substantially affect other metrics (\eg, the mean), and as it also represents the distance for which 50\% of the samples become adversarial.

To avoid a computationally-demanding grid search on the hyperparameter space, \emph{Bayesian Optimization} (BO)~\cite{snoek12_bayes} can be leveraged to build a differentiable approximation of how the objective (\ie, the median perturbation size, in our case) changes as a function of the input hyperparameters. 
Accordingly, gradient descent can be used to efficiently optimize the choice of the best hyperparameters, while improving the approximation of the objective function after each evaluation.

\myparagraph{Bayesian Optimization.}
In our case, BO enables estimating the median perturbation that would be achieved at the end of the FMN algorithm, but without running it for all values.
It requires setting a number of trials \numtrials, which refers to the number of times a new input will be sampled and a new output computed. 
Within \numtrials, an initial number of \numinit trials are used as a ``preliminary'' stage to get a first set of observations to fit an initial model (which approximates the objective). 
Specifically, we first sample a set of \numinit initial hyperparameters (\autoref{line:sobol}) with an external pseudo-random generation process, run the FMN algorithm with them, and collect a set of pairs $(\vct \hyperparameter, \Tilde{\|\vct \delta\|})$ (\autoref{line:fmn_1} - \ref{line:update_set1}).
Then, we fit a Gaussian Process Regression (\texttt{GPR}) model on the observed median over the collected trials (\autoref{line:fit_reg}). 
\texttt{GPR} is a probabilistic model in which multiple regression functions are fitted and averaged, thus reporting, as a function of the value of $\vct \hyperparameter$, a mean and uncertainty value of the metric $\Tilde{\|\vct \delta\|}$.
When fitted, the \texttt{GPR} model predicts $\Tilde{\|\vct \delta\|}$ from a set of hyperparameters $\vct \hyperparameter$.

For the remaining \numtrials - \numinit trials, the BO process involves the definition of an \emph{acquisition function} $a()$, which defines the exploration strategy for navigating the possible hyperparameters.
This function will produce new samplings of $\vct \hyperparameter$ to improve the approximation provided by the \texttt{GPR} model (\autoref{line:acquire}).
Accordingly, the new values of the hyperparameters are chosen where the acquisition function is maximized.
As the acquisition function, we use Noisy Expected Improvement (NEI), which aims to balance exploration (\ie testing new \textit{unexplored} values of the hyperparameter space) and exploitation (\ie refining solutions closer to already seen values) in the search space.
NEI extends the Expected Improvement (EI) criterion.
The EI criterion is defined as the expectation on a candidate of its improvement over the function being estimated:
\begin{equation}
\text{EI}(\vct{x}) = \mathbb{E}[(f(\vct{x}) - f_{\text{min}}) \cdot \mathbb{I}(f(\vct{x}) > f_{\text{min}})] \, ,
\end{equation}
where $f(\vct{x})$ is the objective function to be optimized, $f_{\text{min}}$ is the current best observed value, and $\mathbb{I}(\cdot)$ is the indicator function.
In the presence of noise, directly evaluating $f(\mathbf{x})$ can be unreliable. 
Therefore, NEI incorporates this uncertainty by modeling the noise in the objective function~\cite{letham2019constrained}.

As new hyperparameters are evaluated, new $(\vct \hyperparameter, \Tilde{\|\vct \delta\|})$ pairs are collected, thus the \texttt{GPR} model is updated for an improved estimate (\autoref{line:fmn2} - \ref{line:update_set2}).

At each iteration, the algorithm tracks the best median found (and the corresponding hyperparameters) in order to return the best solution (\autoref{line:store_best_hypers}).

The process is repeated over each of the $N$ configurations to iteratively improve the approximation and find the best set of hyperparameters $(\vct \hyperparameter_1^\star, \ldots, \vct \hyperparameter_N^\star)$, which are returned at the end of the algorithm (\autoref{line:return_hyper}).

\begin{figure}
\centering
  \centering
  \includegraphics[width=\linewidth, scale=2]{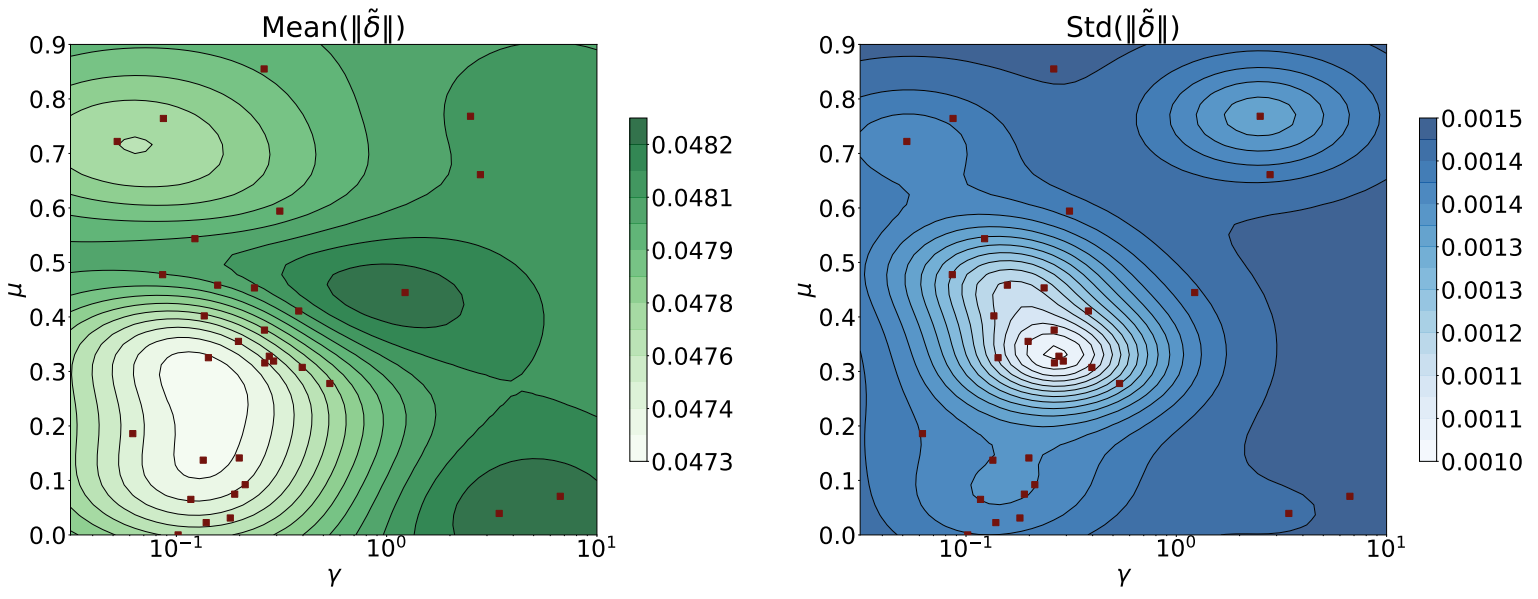}
\caption{
Mean and standard deviation of the median perturbation size $\|\Tilde{\vct \delta}\|$ estimated by the \texttt{GPR} model, for a specific test configuration, as a function of the learning rate~(\learningrate) and momentum (\momentum) hyperparameters. The pairs (\learningrate, \momentum) sampled during the process to iteratively refine the \texttt{GPR} model are shown as red points.}
\label{fig:mean_std_lr_wd}
\end{figure}

As an example of the BO process, we show in  \autoref{fig:mean_std_lr_wd} a GP regressor, isolated on the learning rate (\learningrate) and momentum (\momentum) hyperparameters of a \sgd optimizer.\footnote{ 
Although the sampling of hyperparameters involves an entire set, we isolate it to create a 2-D visualization.}
The plot shows the mean (left) and standard deviation (right) of the estimated $\Tilde{\|\vct \delta\|}$ for each pair of hyperparameters, obtained with $\numtrials=32$ trials. 
Based on the uncertainty estimate, we notice how the acquisition function has focused on exploitation when sampling small, close-to-each-other learning rates, as the uncertainty grows for growing learning rates (indicating sporadic sampling of higher values).

\section{Experimental Analysis}\label{sec:experiments}
In this section, we present the experimental details and results of our proposed optimization framework. 
HO-FMN, given a set of configurations \configurations and a model $\model_\modelindex$ parameterized by $\vct \theta_\modelindex$, finds the best configuration \configuration (\autoref{sec:ho-fmn_configs}) for which it identifies the best set of hyperparameters $\hyperparameter^\star$ (\autoref{sec:ho-fmn_optim}) on which to run the attack. 
Therefore, we first detail the general experimental setting details (\autoref{sec:exp_settings}). 
Then, we describe the creation of the configurations, the hyperparameters associated with each configuration, and the results (\autoref{sect:ho_results}). 
To validate the benefits of our approach, after finding the best hyperparameters for each configuration-model pair, we run the FMN attacks and compare them with other competing attacks, as well as with the FMN baseline (\autoref{sect:attack_results}).
Additionally, we conduct a study on the computational overhead of our attack, showing that our proposed method is the best trade-off between runtime and completeness of the evaluation.

\subsection{Experimental Settings}\label{sec:exp_settings}
We list here the main experimental details employed throughout both hyperparameter optimization and attack runs. We implement HO-FMN in PyTorch, and we run all experiments in a workstation equipped with an NVIDIA RTX A6000 GPU 48 GB. We implement our Bayesian Optimization (BO) search with the Ax framework.\footnote{\url{https://ax.dev//versions/0.1.2/index.html}}

\myparagraph{Datasets.} We take a subset of $4096$ samples ($32$ batches of $128$ samples each) from the CIFAR-10 test set. Instead, for the ImageNet dataset, we selected $1000$ samples to show how the framework can be extended and scaled.
These samples serve to optimize HO-FMN, thus it can be seen as a training set.
Then, for testing the capabilities of the optimized HO-FMN attack, we run the attacks with the best configurations on a separate set of $1000$ samples for both the CIFAR-10 and ImageNet test set. 

\myparagraph{Perturbation Model.} We restrict our analysis to the $\ell_\infty$-norm perturbation model, as it is widely used in SoA evasion attacks and benchmarks~\cite{croce2020robustbench}.
We also point out that, in the original paper, the $\ell_\infty$ perturbation model is observed to be the most challenging for FMN~\cite{pintor2021fast}.

\myparagraph{Models.} We consider 12 state-of-the-art robust models from the RobustBench repository~\cite{croce2020robustbench}, denoted as \textit{$\model_1$}-\textit{$\model_{12}$}. We aim to verify the effectiveness of our method, with a wide range of robust models. The first 9 models are trained for robustness on the CIFAR-10 dataset on a perturbation budget of $\epsilon=8/255$; the remaining 3 models are trained for robustness on ImageNet, within a perturbation budget of $\epsilon=4/255$.
\textit{$\model_1$},\textit{$\model_2$}~\cite{wang23-better}, a WideResNet-70-16 and a WideResNet-28-10 respectively, leverage an improved denoising diffusion probabilistic model (DDPM) to enhance adversarial training.
\textit{$\model_3$},\textit{$\model_5$},\textit{$\model_9$}~\cite{gowal21-improving}, a WideResNet-70-16, a WideResNet-28-10 and a  ResNet-18, use generative models to synthetically expand the original dataset and improve model resilience against $\ell_p$ norm attacks. \textit{$\model_4$}~\cite{rebuffi22-fixing}, a WideResNet-106-16, use a combination of heuristics-based data augmentations and model weight averaging to improve the model's robustness. 
\textit{$\model_6$}, \textit{$\model_8$}~\cite{pang22-robustness}, respectively a WideResNet-70-16 and a WideResNet-28-10, use a self-consistent robust error measure to balance robustness and accuracy. \textit{$\model_7$}~\cite{sehwag22-proxy}, a ResNet-152, uses proxy distributions from diffusion models to enhance adversarial training. Finally, \textit{$\model_{10}$}, \textit{$\model_{11}$}, and \textit{$\model_{12}$}, are transformer architectures adversarially trained using $\ell_{\infty}$ norm perturbation bounded at $\epsilon = 4/255$. The models \textit{$\model_{10}$}, \textit{$\model_{11}$} are two Swin-L and ConvNeXt-L models~\cite{liu2023comprehensive}, while \textit{$\model_{12}$} is a ConvNeXt-L + ConvStem~\cite{singh2023revisiting}.

\myparagraph{Performance Metrics.} 
Within the optimization framework, we employ the smallest median perturbation $\Tilde{\|\vct \delta\|}$ as a criterion to find, for each configuration $\configuration_1, \dots \configuration_N \in \configurations$, the best set of hyperparameters $\vct \hyperparameter^\star$. 
The choice of the median follows the approach employed in the original FMN paper~\cite{pintor2021fast}.

We then rank, based on the resulting $\Tilde{\|\vct \delta\|}$, the configurations $\configuration_1, \dots \configuration_N \in \configurations$.
We take the top-3 configurations, that we name $\configuration_1, \configuration_2,$ and $\configuration_3$ (ordered in terms of performance, lowest median first) to evaluate the robust accuracy at $\epsilon$ ($\robustaccuracy_\epsilon$) of the models at $\epsilon=8/255$ for CIFAR-10, and at $\epsilon=4/255$ for ImageNet (we denote it directly as $\robustaccuracy$ in the rest of the section), following the RobustBench benchmark~\cite{croce2020robustbench}.
Moreover, as explained in \autoref{sec:fmn}, the benefit of FMN is that we can obtain $\robustaccuracy$ by counting the successful attack samples that achieve misclassification with a perturbation size $\|\vct \delta\|_\infty \leq \epsilon$, but we also get, with the same computational cost, the robustness evaluation curve~\cite{biggio18}.

\myparagraph{Search Space for the Configurations.} We first present the experimental settings for the hyperparameter optimization step. 
We list the configurations created from each loss \loss, optimizer \optimizer, and scheduler \scheduler, detailing the sets \losses, \optimizers, and \schedulers respectively. 
As introduced in \autoref{sec:ho-fmn}, we generalize the algorithm to use a selection of:
(i) the loss function $\loss$, selecting between the logit loss (\logitloss)~\cite{carlini2017towards}, the cross-entropy loss (\ce), and the difference of logits ratio (\dlr)~\cite{croce2020autoattack}; (ii) the optimizer \optimizer, selecting between Gradient Descent (\sgd), Adam (\adam), and AdaMax (\adamax); and (iii) the step-size scheduler, selecting among Cosine Annealing (\calr) and Reduced On Plateau (\rlrop).
We report the details of the loss used in \autoref{tab:loss_functions}.

\begin{table}[t]
\caption{Loss functions used in this work. We use $z_\classindex(\vct x; \vct \theta)$ to denote the softmax-scaled outputs of the model, and the indices $\pi_1, \dots, \pi_\numclasses$ to sort the logits as $f_{\pi_1} \geq \cdots \geq f_{\pi_J}$.}
\label{tab:loss_functions}
\resizebox{\columnwidth}{!}{%
\centering
\begin{tabular}{l|c|c}
\toprule
\textbf{Loss Function} & \textbf{Symbol} & \textbf{Equation} \\ \hline
Cross-Entropy & \ce & $\loss_{\ce}(\vct x, y; \vct \theta) =\log(z_y(\vct x; \vct \theta))$ \\
Logits Difference & \logitloss &  $\loss_{\logitloss}(\vct x, y; \vct \theta) =  f_{y}(\vct x; \vct \theta) - \max_{\classindex \neq y}f_{\classindex}(\vct x; \vct \theta)$ \\ 
Difference of Logit Ratio & \dlr & $\loss_{\dlr}(\vct x, y; \vct \theta) = \frac{f_{y}(\vct x; \vct \theta) - \max_{\classindex \neq c}f_{\classindex}(\vct x; \vct \theta)}{f_{\pi_1}(\vct x; \vct \theta) - f_{\pi_3}(\vct x; \vct \theta)}$ \\
\bottomrule
\end{tabular}
}
\end{table}

As explained in \autoref{sec:fmn}, our reformulation of the FMN algorithm enables the use of components already implemented in existing libraries.
Accordingly, we leverage the existing implementations of the aforementioned losses, optimizers, and schedulers as implemented in the PyTorch library, with a few exceptions.
First, the \logitloss and \dlr losses are not implemented in PyTorch. 
Thus we took the implementations from the original FMN algorithm (\logitloss) and from the AutoAttack repository (\dlr).
\footnote{\url{https://github.com/fra31/auto-attack}.}
In addition, despite being the modular FMN version adaptable to each kind of third-party scheduler compatible with the optimizers, we opt for a modified \rlrop implementation, as the original one adjusts a single learning rate ($\learningrate$) for the entire batch. 
Namely, the original implementation of \rlrop decreases \learningrate when there is no improvement (\ie, on plateau) on the average loss for a batch. 
However, we are interested in having a specific adaptation of the learning rate for each sample separately (as for each sample we are in a different region of the loss landscape, and we consider these optimization processes as independent from each other), thus we require a sample-wise \rlrop algorithm that tracks the improvement over the value of the loss on each sample rather than on the average loss of the batch. 
Therefore, we re-implemented the scheduler to have sample-wise control over the learning rates. 
Specifically, for a batch, we are seeking a vector of learning rates $\vct \learningrate$ with one value for each sample in the batch. 
We configure a weighting vector initialized as $\vct w = \vct 1$ containing one weight for each sample of the batch, and we obtain the learning rate by multiplying the weighting vector $\vct w$ for the initial learning rate $\learningrate_0$, \ie, $\vct \learningrate = \vct w \learningrate_0$.
Subsequently, we track the individual loss for each sample, and we multiply by a reducing factor ($<1$) the weight of the weighting vector $w_i$ if the metric stops improving for sample $i$ of the batch over a given number of iterations (as the \textit{patience} parameter of \rlrop).

While \sgd is associated with a scheduler, \adam and \adamax present an inner scheduling procedure, so we fix the scheduler to \fixed, \ie no scheduler, in this case. 
Therefore, for each model, instead of having $|\losses|=3$, $|\optimizers|=3$, and $|\schedulers|=2$, for a total of $N=18$ configurations, we reduce to $N=12$ configurations.

\myparagraph{Hyperparameters.}
We now list the set of hyperparameters \hyperparameters associated to each configuration.
As explained in \autoref{sec:ho-fmn_optim}, this induces a second level on the search space that is different for each \configuration.
Specifically, each optimizer \optimizer and each scheduler \scheduler comes with their own hyperparameters (respectively, $\vct \hyperparameter_\optimizer$ and $\vct \hyperparameter_\scheduler$). 
To reduce the search space, we fix some of the hyperparameters that we denote as ``fixed''.
For the others, we define either the search ranges (Range) and, when different than linear, the scale used for the uniform sampling, or the possible choices (Choice).
We list the hyperparameters search space \hyperparameters, along with their sampling options, in \autoref{tab:opt_sch_params}. 
\begin{table}[h]
\caption{List of the chosen hyperparameters for each Optimizer and Scheduler selected for HO-FMN. As Optimizers, we chose \sgd, the one from the original FMN implementation, and \adam/\adamax; the first requires a Scheduler while the others have an auto-scheduling mechanism. As Schedulers, we selected \calr and \rlrop (our sample-wise implementation). The hyperparameters can be \texttt{range}, \texttt{choice}, or \texttt{fixed}; the sampling distribution can be \texttt{uniform} (default) or \texttt{logarithmic} for better exploring higher ranges. The Optimizers have the most configurable setting, resulting in a larger search space. \\
\scriptsize{($^\star$) The \rlrop scheduler implements our sample-wise version, so the batch size parameter is removed.}}
\label{tab:opt_sch_params}
\centering
\resizebox{\columnwidth}{!}{%
\renewcommand{\arraystretch}{1.2}
\begin{tabular}{l|l|l}
\toprule
\textbf{Optimizer} & \textbf{Hyperparameter} & \textbf{Search Space} \\
\hline
\multirow{4}{*}{\sgd} & learning rate ($\gamma$) & \texttt{range}: [$8/255$,$10$] \texttt{logarithmic} \\
    \cline{2-3}
     & momentum ($\mu$)& \texttt{range}: [$0.0$, $0.9$] \\
    \cline{2-3}
     & weight decay ($\lambda$) & \texttt{range}: [$0.01$, $1.0$] \\
    \cline{2-3}
     & dampening ($\tau$)& \texttt{range}: [$0.0$, $0.2$] \\
    \hline
\multirow{4}{*}{\adam/\adamax} & learning rate ($\gamma$) & \texttt{range}: [$8/255$,$10$ ] \texttt{logarithmic} \\
    \cline{2-3}
     & weight decay ($\lambda$)& \texttt{range}: [$0.01$, $1.0$] \\
    \cline{2-3}
     & eps & \texttt{fixed}: $1e-8$ \\
    \cline{2-3}
     & betas ( $\beta_1$, $\beta_2$) & \texttt{range}: [$0.0$, $0.999$] \\ 
 
\hline
    {\textbf{Scheduler}} & 
    {\textbf{Hyperparameter}} & 
    {\textbf{Search Space}} \\
\hline

\multirow{3}{*}{\calr} & T\_max &  \texttt{fixed}: \steps \\
    \cline{2-3}
     & eta\_min & \texttt{fixed}: $0$ \\
    \cline{2-3}
     & last\_epoch & \texttt{fixed}: $-1$ \\
    \hline
\multirow{5}{*}{\rlrop} & batch size & \texttt{fixed}: $-^\star$ \\
    \cline{2-3}
     & factor  & \texttt{range}: [$0.1$, $0.5$] \\
    \cline{2-3}
     & patience & \texttt{choice}: [$2$, $5$, $10$] \\
    \cline{2-3}
     & threshold & \texttt{fixed}: $1e-4$ \\
     \cline{2-3}
     & eps & \texttt{fixed}: $1e-8$ \\
 \bottomrule
\end{tabular}%
}
\end{table}

\myparagraph{Ax Framework Configuration.} 
The Ax framework works by instantiating multiple trials sequentially. 
Specifically, we employed $\numtrials=32$ trials, of which the first initialization set, \ie $\numinit=8$, are quasi-randomly generated (using the SOBOL~\cite{Bossek_2020} approach), and the remaining $24$ are sampled from the regression models, as implemented by the BOTORCH algorithm~\cite{balandat2020botorch}.

\subsection{Hyperparameter Optimization Results}\label{sect:ho_results}
In this section, we present the results of the hyperparameter optimization. 
We first considered each pair of configurations and models $(\configuration_i, \model_\modelindex)$. 
Then, we tuned each configuration, finding the set of best hyperparameters $\vct \hyperparameter^\star$ that achieve the smallest median perturbation. 
We ranked the configurations by $\Tilde{\|\vct \delta\|}$, and we selected the top-$3$ for each model.

\myparagraph{Tuning Results.} 
In \autoref{tab:best_configs}, we show the resulting top-$3$ configurations that achieve the smallest $\Tilde{\|\vct \delta\|}$ for each model on CIFAR-10, and in \autoref{tab:best_configs_imagenet}, we present the corresponding top-$3$ configurations for ImageNet. 
We highlight how the \dlr loss consistently finds better perturbations than \ce and \logitloss. 
In addition, we found that the \sgd-\calr-\dlr configuration, ranked in the top-$3$ for each model, is also the best one in $6$ over $9$ models.
It's worth noting that this configuration is also very close to the one used by the original FMN, though changing the loss from \logitloss to \dlr and having an optimized set of $\hyperparameter$. 
This loss works well across models as the normalization at the denominator makes the loss (and gradient) more scale invariant (compared to \logitloss that can change of orders of magnitude from one model to the other), easing the tuning of the step size and of the other parameters for the optimization.

\subsection{Best Attacks}\label{sect:attack_results}

\begin{table*}[h]
\caption{Top-$3$ configurations after the hyperparameter optimization on each model ($M_1$-$M_9$), along with the resulting median perturbation, \ie $\Tilde{\|\vct \delta\|}$, on samples from the CIFAR-10 dataset. 
Then, in order, we show the learning rate ($\gamma$) and weight decay ($\lambda$), the beta coefficients ($\beta_{1,2}$) for \adam/\adamax, and the momentum ($\mu$) and dampening ($\tau$) for \sgd. 
Finally, the last columns indicate the factor (fact.) and patience (pat.) for\rlrop.}\label{tab:best_configs}
\resizebox{\textwidth}{!}{
\begin{tabular}{ccccS[table-format=1.6]|cccc|c}
\toprule
    \multicolumn{5}{c|}{\textbf{ }} & 
    \multicolumn{4}{c|}{\textbf{OPTIM. ($\vct \hyperparameter_\optimizer$)}} & 
    \multicolumn{1}{c}{\textbf{SCHED. ($\vct \hyperparameter_\scheduler$)}} \\
\textbf{Model} & \textbf{\configuration} & \optimizer + \scheduler  & \textbf{\loss} &  $\Tilde{\|\vct \delta\|}$  & $\gamma$ & $\lambda$ & $\beta_1$, $\beta_2$ & $\mu$,$\tau$ & \textbf{fac./pat.} \\
\midrule
\multirow{3}{*}{ $M_1$}     & $\configuration_1$ 
    &\sgd + \rlrop & \dlr &       \B 0.048066  &  3.1373e-02 &        0.374 &    - &   (0.5842, 0.148) &  (0.345, 2) \\
    & $\configuration_2$ & \adamax &\dlr&       0.048176   &  3.3608e-02 &        0.113 &  (0.491,0.868) & - &-\\
    & $\configuration_3$ & \sgd + \calr &\dlr  &       0.048403   &  7.3636e-02 &        0.667 &   - &  (0.3720,0.029) &      -  \\
    \hline 
\multirow{3}{*}{ $M_2$} 
    & $\configuration_1$
    & \sgd + \calr& \dlr &       \B 0.048021      &  8.1149e-02 &         0.683 & - &  (0.0744,0.045) &      - \\
    & $\configuration_2$ & \adam &\dlr&       0.048787     &  6.6351e-02 &         0.433 &  (0.412,0.000) & - & - \\
    & $\configuration_3$
    & \sgd + \rlrop& \dlr &       0.048801  &  4.2894e-02 &         0.111 &  - &  (0.2158,0.100) &  (0.160,2)\\
    \hline{}
\multirow{3}{*}{$M_3$} 
    & $\configuration_1$
    & \adam &\dlr &       \B 0.050735    &  8.8306e-02 &        0.577 &  (0.688,0.713) & - &-\\
    & $\configuration_2$
    & \sgd + \calr& \dlr &       0.050961    &  1.9881e-01 &        0.170 & - &  (0.1298, 0.173) & -  \\
    & $\configuration_3$
    & \adamax &\dlr&       0.052110  &  3.1373e-02 &        0.982 &  (0.362,0.751) & - &-  \\
    \hline{}
\multirow{3}{*}{$M_4$} 
    & $\configuration_1$
    & \adam &\dlr  &       \B 0.051502   &  3.1373e-02 &        0.435 &  (0.221,0.816) & - &- \\
    & $\configuration_2$
    & \sgd + \calr &\logitloss   &    0.051720   &  6.2728e-02 &        0.676 &  - &  (0.4512,0.149) &     -    \\
    & $\configuration_3$
    & \sgd + \calr &\dlr &       0.051725     &  3.1373e-02 &        0.924 &  - &   (0.4195,0.130) &    - \\
    \hline{}
\multirow{3}{*}{$M_5$} 
    & $\configuration_1$
    & \sgd + \calr &\dlr   &       \B 0.051760    &  2.9909e-01 &        0.010 & - &  (0.2493,0.105) &  -\\
    & $\configuration_2$
    & \sgd + \calr &\ce &       0.051958      &  3.8703e-02 &        0.511 & - &  (0.3857,0.074) &  - \\
    & $\configuration_3$
    & \adam&\dlr  &       0.052237  &  3.1373e-02 &        0.697 &  (0.275,0.137) & - & -  \\
    \hline{}
\multirow{3}{*}{$M_6$} 
    & $\configuration_1$
    & \sgd + \calr &\dlr  &       \B 0.047542   &  7.6798e-02 &        0.747 & - &  (0.4471,0.091) &  -\\
    & $\configuration_2$
    & \adam &\dlr &       0.047696    &  9.2127e-02 &        0.596 &  (0.687,0.286) & - &- \\
    & $\configuration_3$
    & \adamax &\dlr &       0.047820 &  8.8301e-02 &        0.279 &  (0.264,0.999) & - &-  \\
    \hline{}
\multirow{3}{*}{$M_7$} 
    & $\configuration_1$
    & \sgd + \calr& \dlr &       \B 0.049981    &  6.1492e-02 &        0.061 & - &  (0.3780,0.041) &-\\
    & $\configuration_2$
    & \sgd + \calr &\logitloss &       0.050134     &  6.8632e-02 &        1.000 & - &  (0.1610,0.200) &-\\
    & $\configuration_3$
    & \adam &\dlr   &       0.050165  &  4.9743e-02 &        0.818 &  (0.622,0.255) & - &-   \\
    \hline{}
\multirow{3}{*}{$M_8$} 
    & $\configuration_1$
    & \sgd + \calr &\dlr &       \B 0.045454    &  2.9001e-01 &        0.010 & - &  (0.3191,0.097) & -\\
    & $\configuration_2$
    & \adamax &\dlr &       0.046265 &  5.4455e-02 &        0.248 &  (0.446,0.568) & - &-  \\
    & $\configuration_3$
    & \sgd + \rlrop &\dlr &        0.046554 &  5.1549e-02 &        0.775 & - &  (0.8750,0.078) &  (0.324,5) \\
    \hline{}
\multirow{3}{*}{$M_9$} 
    & $\configuration_1$
    & \sgd + \calr &\dlr  &       \B 0.043584   &  5.3307e-02 &        0.613 & - &  (0.7285,0.165) & -\\
    & $\configuration_2$
    & \sgd + \calr &\logitloss &       0.043850     &  1.8606e-01 &        1.000 & - &  (0.0,0.200) &- \\
    & $\configuration_3$
    & \adam & \ce   &       0.044207    &  4.5904e-02 &        0.456 &  (0.104,0.496) & - &- \\
    \bottomrule
    
\end{tabular}
\hspace{\fill}%
}
\end{table*}

\begin{table*}[h]
\caption{Top-$3$ configurations for $M_{10}$-$M_{12}$, along with the resulting median perturbation, on samples from the ImageNet dataset. 
For further details please refer to \autoref{tab:best_configs}.}\label{tab:best_configs_imagenet}
\resizebox{\textwidth}{!}{
\begin{tabular}{ccccS[table-format=1.6]|cccc|c}
\toprule
    \multicolumn{5}{c|}{\textbf{ }} & 
    \multicolumn{4}{c|}{\textbf{OPTIM. ($\vct \hyperparameter_\optimizer$)}} & 
    \multicolumn{1}{c}{\textbf{SCHED. ($\vct \hyperparameter_\scheduler$)}} \\
\textbf{Model} & \textbf{\configuration} & \optimizer + \scheduler  & \textbf{\loss} &  $\Tilde{\|\vct \delta\|}$  & $\gamma$ & $\lambda$ & $\beta_1$, $\beta_2$ & $\mu$,$\tau$ & \textbf{fac./pat.} \\
\midrule
\multirow{3}{*}{ $M_{10}$}     & $\configuration_1$ 
    &\sgd + \calr & \logitloss &       \B 0.016130 &  8.2874e-02 &        0.266 &    (0.4962, 0.037) &   - &  - \\
    & $\configuration_2$ & \adamax &\dlr&       0.017020 &  8.1110e-01 &        0.404 &  (0.566,0.098) & - &-\\
    & $\configuration_3$ & \sgd + \calr &\dlr  &       0.017138
   &  3.1036e-01 &        0.755 &   - &  (0.8011,0.055) &      -  \\
    \hline 
\multirow{3}{*}{ $M_{11}$} 
    & $\configuration_1$
    & \adamax & \dlr &       \B 0.012425
      &  1.9174e-01 &         0.304 & (0.387,0.367) &  -&      - \\
    & $\configuration_2$ & \adam &\dlr&       0.013496     &  9.4361e-02 &         0.244 &  (0.354,0.231) & - & - \\
    & $\configuration_3$
    & \sgd + \rlrop& \dlr &       0.015658 &   2.1192e-02 &          0.109 &  - &  (0.1497, 0.155) &  (0.207,5)\\
    \hline{}
\multirow{3}{*}{$M_{12}$} 
    & $\configuration_1$
    & \sgd + \calr &\logitloss &       \B 0.014309    &  1.0000e+01 &        0.582 &  ( 0.9000,0.200) & - &-\\
    & $\configuration_2$
    & \adamax & \dlr &       0.014430    &  5.8146e-01 &        0.401 & - &   (0.256,0.332) & -  \\
    & $\configuration_3$
    & \adam &\dlr&       0.014899  &  8.0030e+00 &        0.568 &  -&  (0.806,0.222) &-  \\
    \bottomrule
    
\end{tabular}
\hspace{\fill}%
}
\end{table*}

Given the top-$3$ configurations for each model, on the CIFAR-10 test set \autoref{tab:best_configs} and on the ImageNet test set \autoref{tab:best_configs_imagenet}, we run the final attack evaluation on our test sets, respectively $1000$ samples from CIFAR-10/ImageNet, and compare with the baseline FMN version~\cite{pintor2021fast} to clearly show the benefits of HO-FMN.
In addition, to rigorously validate with SoA, parameter-free approaches, we compare our attack configurations achieved through HO-FMN with the APGD attack in its \ce and \dlr loss versions~\cite{croce2020autoattack}.
We specify that comparing head-to-head the entire AA ensemble against a single attack from HO-FMN, would result in a four-versus-one evaluation, ultimately indicating a disproportionate analysis.
Finally, to let the comparison be as fair as possible, we ensure that all algorithms are initialized in the same way.
Specifically, we ensure that APGD does not perform Expectation over Transformations (EoT) steps, \ie, the computation of a smoothed gradient before the actual attack loop and the restarts. 
We removed this step as we want to avoid the variability given by the randomness in the EoT procedure. Thus, by avoiding this particular initialization, we can ensure that all attacks start from the same $\vct x$ and have no advantage (or disadvantage) given by random initializations of $\vct \delta_0$.
We remark that the same method is not computed in the default configuration of APGD. Additionally, this initial EoT can also be seamlessly added later to FMN and HO-FMN.

\begin{table}[t]
\caption{Robust Accuracy (RA) with fixed perturbation $\epsilon$=8/255 computed for each model $M_1-M_9$ with, respectively, the \textbf{Baseline} FMN attack, the two \textbf{APGD$_{\textbf{\ce/\dlr}}$} versions and the top-3 HO-FMN configurations of each model ($C_1$, $C_2$, $C_3$). Except for two models, we beat both baseline and APGD attacks.}
\label{tab:asr_table}
\resizebox{\columnwidth}{!}{%
\centering
\setlength{\tabcolsep}{4pt} 
\renewcommand{\arraystretch}{0.7}
\begin{tabular}{llc|llc|llc}
\toprule
\textbf{Model}&\textbf{Attack}&\textbf{\robustaccuracy} & \textbf{Model}&\textbf{Attack}&\textbf{\robustaccuracy} & \textbf{Model}&\textbf{Attack}&\textbf{\robustaccuracy} \\
\hline
\multirow{6}{*}{$M_1$} &  Baseline &   0.744  & \multirow{6}{*}{$M_2$} &  Baseline &   0.716 & \multirow{6}{*}{$M_3$}&  Baseline &   0.704 \\
&  APGD$_{\dlr}$ &   0.718 &  &  APGD$_{\dlr}$ &   0.687&  & APGD$_{\dlr}$ &   0.684 \\
&   APGD$_{\ce}$ &   0.741 &   & APGD$_{\ce}$ &   0.716 &   & APGD$_{\ce}$ &   0.687 \\
&       $C_1$ & 0.724   &       & $C_1$ &  0.688  & &$C_1$ &   0.683 \\
&       $C_2$ &   0.718 &       & $C_2$ &  \B{0.683}  &  &$C_2$  &  \B{0.678} \\
&       $C_3$ &   \B{0.717} &       & $C_3$ &   0.693 &   &$C_3$ &  0.681  \\
\hline
\multirow{6}{*}{$M_4$} &  Baseline &   0.680 & \multirow{6}{*}{$M_5$}&  Baseline &   0.679 & \multirow{6}{*}{$M_6$} &  Baseline &   0.664 \\
&  APGD$_{\dlr}$ &   0.661&  &APGD$_{\dlr}$ &   0.659 &  &APGD$_{\dlr}$ &   \B{0.631}\\
&   APGD$_{\ce}$ &   0.678 &   &APGD$_{\ce}$ &   0.656 &   &APGD$_{\ce}$ &   0.658\\
&       $C_1$ &   0.661 &       &$C_1$ &   \B{0.652}&       &$C_1$ &  0.633 \\
&       $C_2$ &  0.661  &       &$C_2$ &   0.658&       &$C_2$ &   0.637  \\
&       $C_3$ &  \B{0.657}   &       &$C_3$ &   \B{0.652}&     &$C_3$ &    0.638  \\
\hline
\multirow{6}{*}{$M_7$} &  Baseline &   0.672 &\multirow{6}{*}{$M_8$} &  Baseline &   0.639 &  \multirow{6}{*}{$M_9$} &  Baseline &   0.635\\
&  APGD$_{\dlr}$ &   0.647 &  &APGD$_{\dlr}$ &   \B{0.616} &  &APGD$_{\dlr}$ &   0.616 \\
&   APGD$_{\ce}$&   0.654 &   &APGD$_{\ce}$ &   0.651 &   &APGD$_{\ce}$ &   0.610\\
&       $C_1$ &   \B{0.638} &   &    $C_1$ & 0.618  & &      $C_1$ &  \B{0.596} \\
&       $C_2$ &   0.641  &   &    $C_2$ &   0.621  & &      $C_2$ & 0.609  \\
&       $C_3$ &  \B{0.638} &      & $C_3$ &   0.624  &   &    $C_3$ & 0.616  \\
\bottomrule
\end{tabular}
}
\end{table}

\begin{table}[t]
\caption{Robust Accuracy (RA) with fixed perturbation $\epsilon$=4/255 computed for each model $M_{10}-M_{12}$ on the ImageNet dataset. We report the numerical results for, respectively, the \textbf{Baseline} FMN attack, the two \textbf{APGD$_{\textbf{\ce/\dlr}}$} versions and the top-1 HO-FMN configuration $C_1$ of each model. For all the models, we beat both baseline and APGD attacks.}
\label{tab:asr_table_imagenet}
\resizebox{\columnwidth}{!}{%
\centering
\setlength{\tabcolsep}{4pt} 
\renewcommand{\arraystretch}{0.7}
\begin{tabular}{llc|llc|llc}
\toprule
\textbf{Model}&\textbf{Attack}&\textbf{\robustaccuracy} & \textbf{Model}&\textbf{Attack}&\textbf{\robustaccuracy} & \textbf{Model}&\textbf{Attack}&\textbf{\robustaccuracy} \\
\hline
\multirow{4}{*}{$M_{10}$} &  Baseline &   0.619  & \multirow{4}{*}{$M_{11}$} &  Baseline &   0.619 & \multirow{4}{*}{$M_{12}$}&  Baseline &   0.614 \\
&  APGD$_{\dlr}$ &   0.611 &  &  APGD$_{\dlr}$ &   0.614&  & APGD$_{\dlr}$ &   0.609 \\
&   APGD$_{\ce}$ &   0.608 &   & APGD$_{\ce}$ &   0.605 &   & APGD$_{\ce}$ &   0.594 \\
&       $C_1$ & \B{0.597}   &       & $C_1$ &  \B{0.600}  & &$C_1$ &   \B{0.588} \\
\bottomrule
\end{tabular}
}
\end{table}

\begin{figure*}[!h]
\centering
\includegraphics[width=0.99\textwidth]{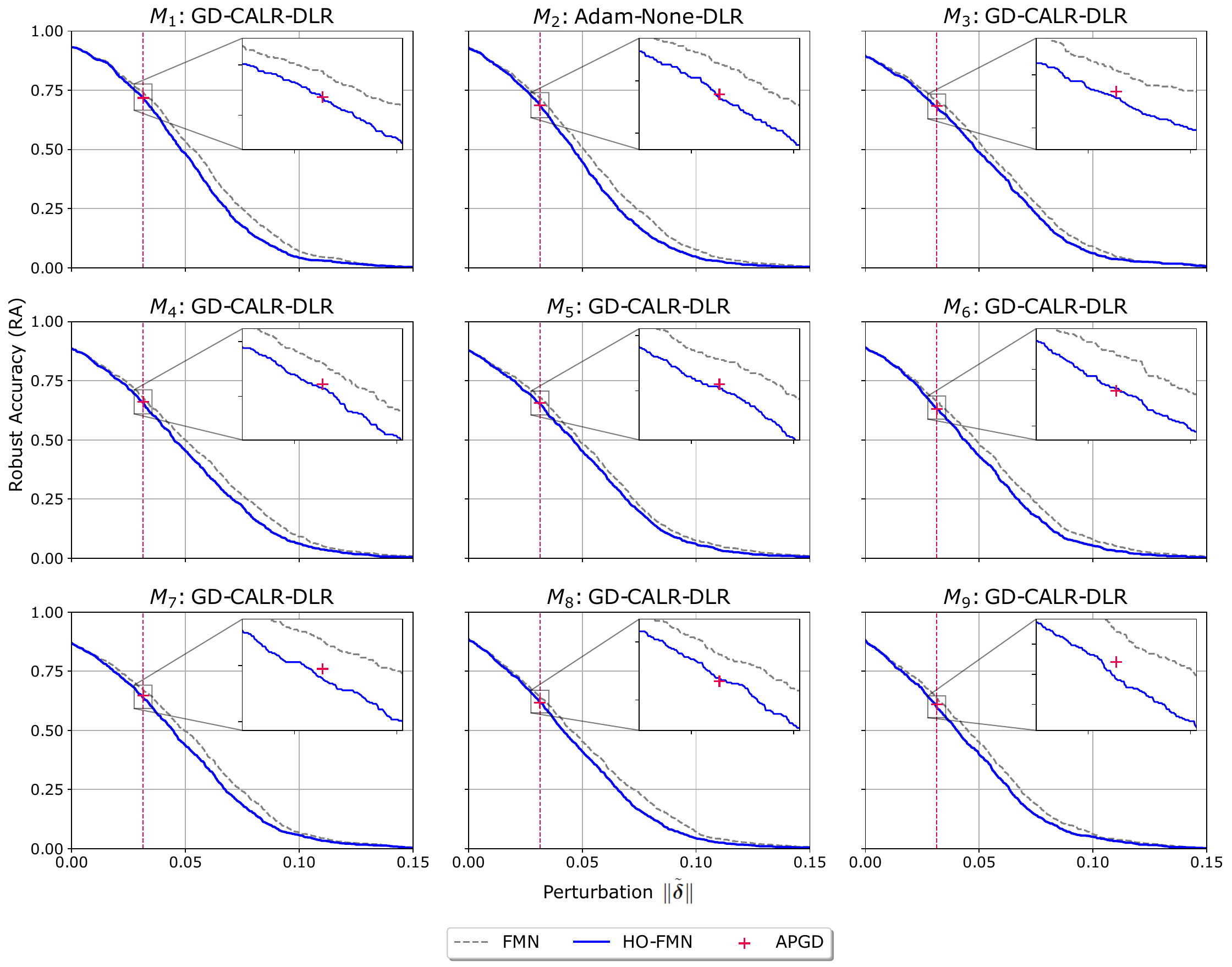}
    \caption{Robustness evaluation curves for $\model_1$-$\model_9$. The dashed-gray and solid-blue lines represent FMN and HO-FMN. The robust accuracy (\robustaccuracy) value at $\epsilon = 8/255$ computed with APGD$_{\ce/\dlr}$ (the best value between the two) is also shown as a red cross.  }\label{fig:sec_evals_cifar10}
\end{figure*}

\begin{figure*}[!h]
\centering
    \includegraphics[width=0.99\textwidth]{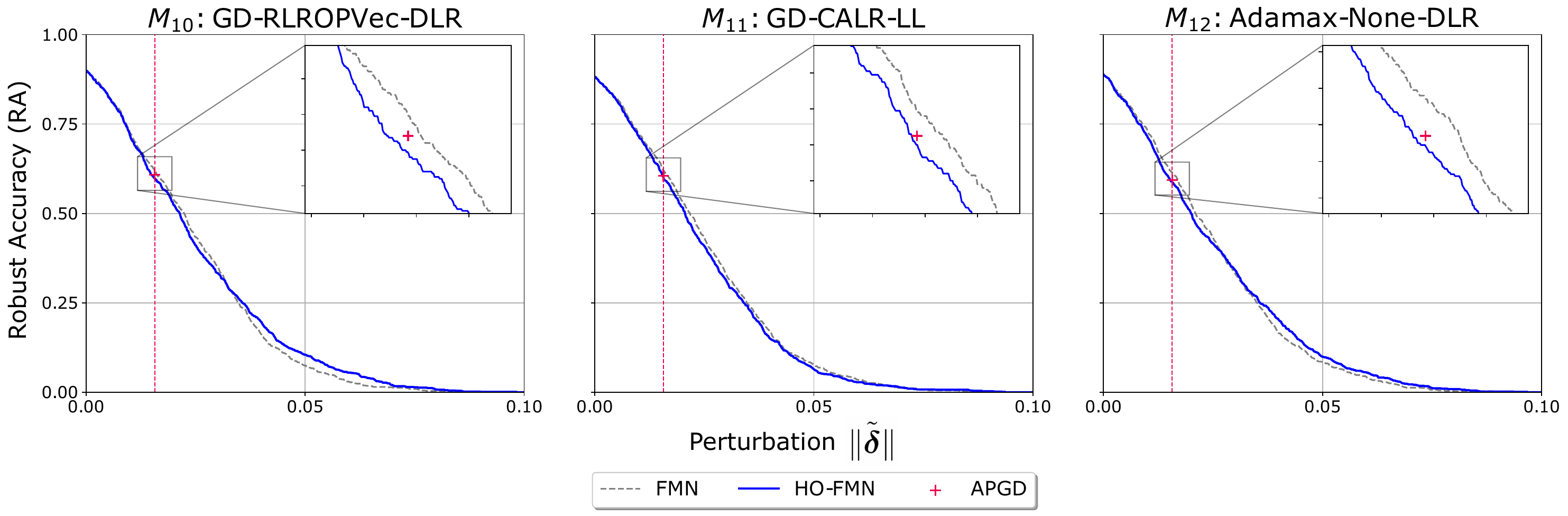}
    \caption{Robustness evaluation curves for $\model_{10}$-$\model_{12}$, and APGD robust accuracy at $\epsilon = 4/255$. Please refer to \autoref{fig:sec_evals_cifar10} for further details.
    }\label{fig:sec_evals_imgnet}
\end{figure*}

\myparagraph{Attack Results.} We show the resulting robustness evaluation curves in \autoref{fig:sec_evals_cifar10} for the CIFAR-10 experiments; while in \autoref{fig:sec_evals_imgnet} we report the curves for the attacks against ImageNet models. 
We compare the curves of the baseline FMN attack against HO-FMN.
The FMN baseline is defined as the original formulation in~\cite{pintor2021fast}, thus configured with \sgd-\calr\logitloss, and optimizer with \learningrate$=1.0$ and \momentum$=0.0$.
In addition, we highlight, for the single perturbation norm of $\|\vct \delta \| = 8 / 255$ (CIFAR-10) and $\|\vct \delta \| = 4 / 255$ (ImageNet), the Robust Accuracy (\robustaccuracy) found by APGD$_{\ce/\dlr}$. 
We selected the attack that performed better, in terms of RA, between the two versions of APGD.
We show the empirical results in \autoref{tab:asr_table} for CIFAR-10 and in \autoref{tab:asr_table_imagenet} for ImageNet. 
In the first case, except for $2/9$ models, HO-FMN outperforms the APGD attack (\ie, the blue line lies below the red cross). In the second case, we are able to beat all the 3 selected models with our HO-FMN version.
Furthermore, our attack computes the robustness evaluation curve with one single run. 
Achieving the same result with APGD is only possible by executing APGD multiple times, as we discuss next.  

\myparagraph{Computational Overhead.}
We perform a set of additional experiments to have a clear understanding of the overhead added by running hyperparameter optimization.
The total time of the HO process is mainly dictated by the time required for a single attack multiplied by the number of trials. Our tuning setting, as described in \autoref{sec:exp_settings}, consists of running 32 trials of HO, each running  FMN on a batch of 128 samples over 200 steps. To analyze the HO overhead added to FMN, we measure the average execution time of a single FMN attack on the same setup and relate it to the number of trials. We refer to this time as $T_{FMN}$, for which we find $T_{FMN} = 7.479$ seconds. We then compute an estimate $\tilde{T}_{HO}$ of the HO process, thus ignoring the Gaussian Processes (GP) overhead, as $\tilde{T}_{HO} = T_{FMN}\cdot32 = 239.328$ seconds. Then, we run the actual HO-FMN under the same setup and measure the total execution time. On average, we find $T_{HO}=262.612$ seconds, which indicates that the difference between our estimate and the measured time equals $\Delta_{HO} = \tilde{T}_{HO}-T_{HO} = 23.284$ seconds. 
Therefore, for a single trial, the required time amounts to $\Delta_{HO} / 32 = 0.727$. 
Compared to $T_{FMN}$, we can assert that the overhead added by the optimization is acceptable in practice.

\begin{table}[t]
\caption{Runtime comparison between HO-FMN (\sgd-\calr-\dlr/\ce) and APGD$_{\ce/\dlr}$ adapted to find a minimum-norm solution (each row represents a binary search iteration). We show the total time and best median $\Tilde{\|\vct \delta\|}$ found by the attack on a batch of 128 samples from CIFAR-10 on model $M_{9}$.}
\label{tab:fmn_apgd_runtime}
\centering
\resizebox{\columnwidth}{!}{%
\begin{tabular}{@{}rrr@{}}
\toprule
                      & \textbf{Total (avg) time {[}s{]}} & \textbf{Best (median) $\Tilde{\|\vct \delta \|}$} \\ \midrule
HO-FMN$_{\ce (\dlr)}$ & \textbf{4.753 (5.257)}            & \textbf{0.053 (0.053)}                         \\
APGD$_{\ce (\dlr)}$ 1 & 3.635 (4.064)                     & 0.062 (0.062)                                  \\
APGD$_{\ce (\dlr)}$ 2 & 7.241 (8.078)                     & 0.062 (0.062)                                  \\
APGD$_{\ce (\dlr)}$ 3 & 10.856 (12.094)                   & 0.062 (0.062)                                  \\
APGD$_{\ce (\dlr)}$ 4 & 14.508 (16.105)                   & 0.054 (0.054)                                  \\
APGD$_{\ce (\dlr)}$ 5 & 18.170 (20.141)                   & 0.054 (0.054)                                  \\ \bottomrule
\end{tabular}
}
\end{table}

\myparagraph{Comparing FMN against APGD.} In \autoref{sec:fmn}, we show how FMN allows us to compute the robustness evaluation curves~\cite{biggio18}, which is instead practically unfeasible for fixed-budget attacks, such as APGD~\cite{croce2020autoattack}. 
Being one of the main advantages of our HO-FMN approach the possibility to create robustness evaluation curves, we conduct additional experiments, summarized in \autoref{tab:fmn_apgd_runtime}, where we measure the required time for HO-FMN (\sgd-\calr-\dlr/\ce averaged) to compute the curves compared to APGD.
In fact, through APGD, it is possible to have only a scalar robustness evaluation: for a given value of $\epsilon$ (i.e., the maximum perturbation that constrains the attack) APGD provides a scalar robust accuracy value associated to the given value $\epsilon$. 
Therefore, to be compared with HO-FMN, we adapted APGD to find a minimum-norm solution using a binary search approach, that we applied sample-wise. 
Within this approach, we define a number of search steps that we set to 5, in addition to the search interval $[\epsilon_{\text{low}}, \epsilon_{\text{high}}]$. 
In particular, $\epsilon_{\text{low}}$ is the lowest value the perturbation budget can take, while $\epsilon_{\text{high}}$ is the highest. 
The binary search algorithm works by selecting a value for the perturbation budget which is always set as $(\epsilon_{\text{high}} - \epsilon_{\text{low}})/2$ (in the middle of the interval), and the interval is updated according to the successfulness of the attack. 
Hence, in the initialization phase, the search interval is set as $[\epsilon_{\text{low}}, \epsilon_{\text{high}}] = [0,32/255]$, and at each step, APGD is run with a perturbation budget of $\epsilon_i=(\epsilon_{\text{high}} - \epsilon_{\text{low}})/2$ ($\epsilon_0$=16/255). 
If the attack finds a successful adversarial perturbation, we narrow the search to the lower half of the interval (i.e., $\epsilon_{\text{high}}=\epsilon_i$); otherwise, we search on the upper half (i.e., $\epsilon_{\text{low}} = \epsilon_i$). 
This process is repeated within the selected half-interval, progressively refining the search until the maximum number of search steps is reached. 
As shown in \autoref{tab:fmn_apgd_runtime}, the first column is the total average time (in seconds) the attack took to complete. We show that our HO-FMN finds the best minimum-norm solution in a single run (first row). 
The next rows, indicated by APGD (i), represent the binary search step i performed by the two APGD versions.  
\autoref{tab:fmn_apgd_runtime} shows
that the best solution for APGD is found at step 4 (APGD (4)), while the binary search continues to step 5 with no improvement. 
Our FMN version, finds its best solution in approximately 5 seconds, while it takes about 20 seconds for each APGD version to complete the binary search, therefore showing the efficacy of our FMN approach in finding the robustness evaluation curve compared to APGD.

\section{Related Work}\label{sec:related}

Adversarial attacks are recognized as an important tool to empirically evaluate the robustness of ML models. 
Many gradient-based attacks have been proposed as an effective tool to assess the models' robustness, and have evolved over time seeking for better efficiency. 
Among the most used attacks, the Projected Gradient Descent (PGD) attack~\cite{madry2017towards} has been extensively used as a bare essential evaluation tool.
However, attacks like PGD, which solve an optimization problem, require a proper hyperparameter configuration (\eg, learning rate, step decay etc.) to avoid suboptimal solutions and, consequently, providing an overestimated adversarial robutness evaluation~\cite{carlini2019evaluating}.
To mitigate this issue, parameter-free approaches that combine multiple attacks~\cite{croce2020autoattack} have also been proposed.

\myparagraph{AutoAttack (AA).} This attack consists of ensembling 4 parameter-free attacks, including Auto-PGD (APGD), \ie, an attack that directly improves the basic PGD optimization by dynamically updating the step size. Together with APGD with both \ce and \dlr losses, AA also uses a gradient-based (Fast Adaptive Boundary~\cite{croce2020minimally} and a black-box (SquareAttack~\cite{andriushchenko2020square}) attack, and ensembles them by retaining the first useful result found by any of them (within the fixed budget), in a sample-wise manner.

\myparagraph{Adaptive Auto-Attack (AAA).} This attack~\cite{yao2020automated} provides a further evolved approach by conceiving the attacks as building blocks, thus having multiple interchangeable parts, and performing an extensive search, virtually constructing a huge ensemble of attacks. 
However, the implemented algorithm does not efficiently filter the searched trials, thus potentially wasting computing resources, and does not optimize the hyperparameters of these attacks, potentially disrupting the evaluation results.

\myparagraph{Limitations of Existing Methods.} Just like standard fixed-epsilon attacks, the robustness evaluation for both AA and AAA are constrained to a single perturbation budget (\eg, $\epsilon = 8/255$), resulting in a scalar robustness estimate. 
Such characteristic of both AA and AAA inhibits them from constructing a full-scale robustness evaluation curve on multiple perturbation values, which would inevitably require multiple attack runs. As also noted in our work, constructing the full curve with these attacks requires running them multiple times to find the smallest $\epsilon$ that satisfies the attack's success.

In this direction, our proposed HO-FMN approach collectively takes advantage of the reliability of the parameter-free paradigm, as well as enabling a thorough robustness evaluation, contrary to the competing parameter-free approaches. 
Our results show the efficacy of the proposed hyperparameter optimization strategy when compared to the baseline FMN attack.

\section{Conclusions and Future Work} \label{sec:conclusions}

In this work, we investigated the use of hyperparameter optimization to improve the performance of the FMN attack. 
Specifically, we reimplemented the FMN attack into a modular version that enables changing the loss, the optimizer, and the step-size scheduler to create multiple configurations of the same attack. We used Bayesian optimization to find the best attack hyperparameters for each configuration selected. Our findings highlight that hyperparameter optimization can improve FMN to reach competitive performance with existing attacks while providing a more thorough adversarial robustness evaluation (\ie, computing the whole robustness evaluation curve).

We argue that the same approach can be combined with other attacks and perturbation models. To this end, we plan to extend our analysis beyond the $\ell_\infty$-norm FMN attack, considering  $\ell_0$, $\ell_1$, and $\ell_2$ norms. 
We remark that adding more hyperparameters to tune would make the search space bigger, resulting in a longer optimization time. To this end, we will also develop sound heuristics to filter the suboptimal configurations without running the full attacks, making hyperparameter tuning more efficient.
Additionally, we will design faster exploration phases in the initial steps of the FMN optimization process to enable further exploration of the loss landscape. 

\section*{Acknowledgments}
This work has been carried out while L. Scionis and G. Piras were enrolled in the Italian National Doctorate on AI run by the Sapienza University of Rome in collaboration with the University of Cagliari; 
and supported by project SERICS (PE00000014) and project FAIR (PE0000013) under the NRRP MUR program funded by the EU - NGEU; the European Union's Horizon Europe Research and Innovation Programme under the project Sec4AI4Sec, grant agreement No 101120393; and Fondazione di Sardegna under the project ``TrustML: Towards Machine Learning that Humans Can Trust’’, CUP: F73C22001320007.

\bibliographystyle{elsarticle-num} 
\bibliography{bibliography}

\end{document}